\documentclass{article}
\usepackage{spconf}
\usepackage{pbox}
\usepackage{times}
\usepackage{latexsym}
\usepackage{color}
\usepackage{times}
\usepackage{bbm}
\usepackage{dsfont}
\usepackage{qtree}
\usepackage{multirow}
\usepackage{url}
\usepackage{latexsym}
\usepackage{graphicx}
\usepackage{times}
\usepackage{url}
\usepackage{latexsym}

\usepackage{times}
\usepackage{url}
\usepackage{latexsym}
\usepackage{amsmath}
\usepackage{amssymb}
\usepackage{amsthm}
\usepackage{color}
\usepackage{graphics}
\usepackage{graphicx}
\usepackage{multirow}
\usepackage{xcolor}
\usepackage{fullpage}

\usepackage{bm}
\usepackage{todonotes}

\title{Speaker-Sensitive Dual Memory Networks for Multi-Turn Slot Tagging}

\name{Young-Bum Kim\textsuperscript{*}, Sungjin Lee\textsuperscript{**},Ruhi Sarikaya\textsuperscript{*}}
\address{Amazon Alexa Brain, Seattle, WA\textsuperscript{*}, \\
Microsoft Research, Redmond, WA\textsuperscript{**}}

\date{}

\begin{document}
\maketitle
\begin{abstract}
In multi-turn dialogs, natural language understanding models can introduce obvious errors by being blind to contextual information. To incorporate dialog history, we present a neural architecture with \emph{Speaker-Sensitive Dual Memory Networks} which encode utterances differently depending on the speaker. This addresses the different extents of information available to the system - the system knows only the surface form of user utterances while it has the exact semantics of system output. We performed experiments on real user data from Microsoft Cortana, a commercial personal assistant. The result showed a significant performance improvement over the state-of-the-art slot tagging models using contextual information.
\end{abstract}

\section{Introduction}
\label{sec:intro}

Major technology companies have been investing in natural language understanding systems as effective tools for human-computer communication~\cite{ybkim2015weakly,kimgazet2015, sarikaya2016overview,ybkim2016reuse,kimfrustratingly,kimdomainless, kim2017domainattention, kim2017advr}. 
Of particular interest are task-oriented personal assistants, e.g., Amazon's Alexa, Apple Siri and Microsoft Cortana. In task-oriented dialogs, the main role of the system is to fill out the associated slots to the semantic frame of the requested task. 
As the demand for fulfilling more complex tasks with voice-interface agents keeps increasing, it becomes crucial to address some issues faced when handling multi-turn interaction.
For example, in Table~\ref{tab:scenarioReservation}, the user's utterance at turn 3 could mean either \emph{number\_people} or \emph{time} without any context.

\begin{table*}[!ht]
\centering
\begin{tabular}{|c|c|c|c|}
\hline
Turn & Speaker & Utterance & Targeted Slots\\ \hline
\multirow{2}{*}{1} & System & Where do you want to reserve? & \emph{place\_name} \\ \cline{2-4}
& User & Pizza Hut & - \\ \hline
\multirow{2}{*}{2} & System & For how many people? & \emph{number\_people} \\  \cline{2-4}
& User & Two & - \\ \hline
\multirow{2}{*}{3} & System & Okay. What time should I make the reservation for? & \{\emph{time, date}\} \\  \cline{2-4}
& User & Actually it's three & - \\ \hline
\end{tabular}
\caption{An example dialog for restaurant reservation. The targeted slots are the slots that the system asks about at each turn.}
\label{tab:scenarioReservation}
\end{table*}

Traditionally, the natural language understanding (NLU) module~\cite{tur2011spoken} delegates the task of contextual interpretation to downstream modules such as dialog state tracker~\cite{williams2016dialog}. However, given that most dialog state trackers just exploit the slot-values recognized by the NLU module as is without further transformation based on context, the NLU errors due to the context-insensitivity are very likely to propagate through downstream modules. Thus it is critical to reduce the context-related NLU errors as early in the system pipeline as possible. 

With this aim, a host of previous studies has attempted to incorporate dialog history for contextual interpretation~\cite{bhargava2013easy,xu2014contextual,shi2015contextual}. A large body of work, however, exploited only previous turn information such as the NLU results at the previous turn. Not only are they limited in the range of history information but also they internally suffer the error propagation problems by directly using previous predictions. With the rise of end-to-end deep neural network models, Chen et al.~\cite{chen2016end} addressed such shortcomings with end-to-end memory networks~\cite{sukhbaatar2015end} where the embeddings of all past user utterances are stored in a memory. In their work, however, no system's information was used which will prove powerful in our experimental results.

In this paper we propose an end-to-end neural architecture for slot tagging that uses \emph{Speaker-Sensitive Dual Memory Networks} (SSDMNs) to leverage both user and system information in a tailored manner. Specifically, since the system only gets to see the surface form of user's utterance, we encode each word of all past user's utterances using the Long Short-Term Memory (LSTM) model~\cite{hochreiter1997long}. We store the encoded vectors of each word in a memory and use an attention mechanism~\cite{bahdanau2014neural} to selectively use only relevant words to the current user input. Whereas the system already knows the exact semantics of its past utterances, thus we directly encode the targeted slots of system's question to avoid parsing noisy natural language expressions. Due to this different nature, we introduce a separate memory storing all system's targeted slot embeddings. The attention mechanism again allows to selectively use relevant system information. Hori et al.~\cite{hori2016context} adopted a similar concept, role-dependent encoding, but their work is not aimed for slot tagging, nor does it exploit the system output semantics. Furthermore, we adopt an efficient pre-training method~\cite{kim2015new,kimpre2015,kim2016scalable, kim2017preframe} for system's targeted slot encoding that further improves the system's performance. To verify the efficacy of the proposed model, we performed evaluation on a large amount of real data gathered from five different domains of commercial personal assistants. In comparison to previous approaches our method showed a large improvement. 

The rest of the paper is structured as follows. In Section~\ref{sec:method} we describe SSDMNs in depth. In Section~\ref{sec:experiments} we discuss our experiments. 
We finish with conclusions in Section~\ref{sec:conclusion}.

\section{Speaker-Sensitive Dual Memory Networks}
\label{sec:method}
At turn $t$, the multi-turn slot tagging model takes a user input ${\bm u}^t$ and outputs semantic tags ${\bm m}^t$ by considering the contextual information, i.e., a sequence of user's past utterances $\{{\bm u}^1, \ldots, {\bm u}^{t-1}\}$ and the corresponding system outputs $\{{\bm m}^1, \ldots, {\bm m}^{t-1}\}$.

\subsection{Overall architecture}
\begin{figure*}
    \centering
    \includegraphics[width=\textwidth]{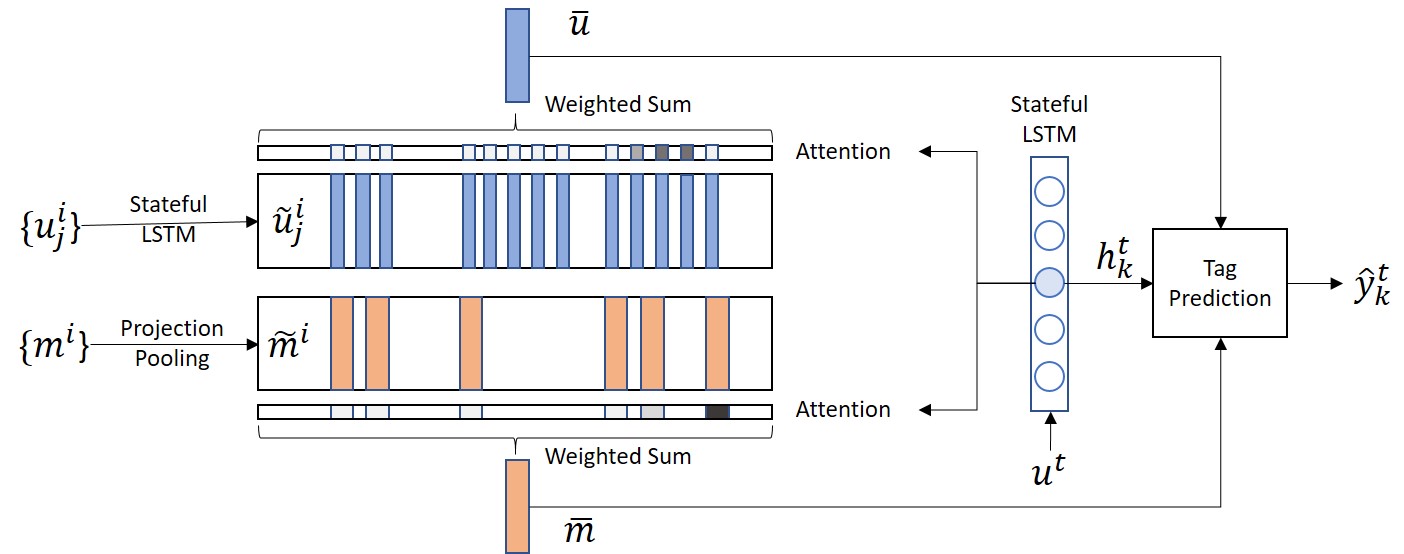}
    \caption{The overall architecture of \emph{Speaker-Sensitive Dual Memory Networks}.}
    \label{fig:overall}
\end{figure*}
The overall SSDMNs architecture is illustrated in Figure~\ref{fig:overall}. 
The model first encodes the last user input and system output to produce vector embeddings. Then it updates the dual memories with the generated vectors. After that, as it encodes the current user utterance, at each word position, it computes the attention weights to find relevant information. Finally it makes prediction for each word by jointly considering the encoded vector of each word and the memory content averaged with respect to the attention weights. The main steps are described below.

\subsection{User utterance encoding} In order to obtain a more contextful vector representation of the $j$-th word $u^i_j$ of the user utterance at turn $i$, we adopt a stateful LSTM instead of a vanilla LSTM encoder - the LSTM output for the last word in an utterance is fed as the initial state into the LSTM network for the next utterance while all LSTM weights are shared. The stateful LSTM runs over both past user utterances and the current input:
$$\tilde{\bm u}^i = \text{StatefulLSTM}({\bm u}^i), i < t$$
$${\bm h}^t = \text{StatefulLSTM}({\bm u}^t)$$

\subsection{System output encoding}
Since the system has the exact semantics of system utterances, we directly encode the targeted slots that usually have strong influence on the following user's response. Concretely, for each system output, we are given an k-hot vector ${\bf m}^i \in \mathbb{R}^l$, where $l$ is the number of slot tags, which indicates which slot is targeted by the system's output. 
Then we use a trainable projection matrix ${\bf P} \in \mathbb{R}^{d \times l}$ to map the k-hot vector to a $d$-dimensional dense vector $\tilde{\bf m}^i$:
$$\tilde{\bf m}^i = {\bf P}{\bf m}^i$$
Note that since ${\bf m}^i$ is a k-hot vector, $\tilde{\bf m}^i$ is the sum of the targeted slot embeddings.
It has been observed that pre-trained embeddings can serve as good initialization values that help avoid converging to bad local optima and speed up model training. Inspired by~\cite{kim2015new}, we adopt \emph{Canonical Correlation Analysis} (CCA) to efficiently pre-train the projection matrix. The CCA algorithm finds a lower dimensional projection that maximizes the correlation between a set of random vector pairs:
\begin{itemize}
\item \{${\bf a}^i$\} is a set of zero vectors in which the entry corresponding to the slot tag of the i-th sentence is set to 1.
\item \{${\bf b}^i$\} is a set of zero vectors in which the entries corresponding to words spanned by the tag are set to 1.
\end{itemize}

\subsection{Memory update}
Although an LSTM network can capture long-term dependencies, it has been argued that LSTM networks' fixed-size memory may not be enough to contain a large amount of information encoded in variable-sized inputs~\cite{bahdanau2014neural}.
Thus before performing the tag prediction we update the dual memories with the encoded vectors of each word of the user utterances ($\tilde{\bf u}^i_k$) and system output vectors ($\tilde{\bf m}^i$) encoded in the manner described above, respectively.

\subsection{Attention mechanism}
Since not every piece of history information is helpful in making prediction for a current word, we selectively use only relevant information.
To this end, we first compute the attention weights for both memories:
$${\bf \alpha}^i_j = \mathrm{softmax} ({\bf W}_\alpha(\tilde{\bf u}^i_j \circ {\bf h}^t_k) + b_\alpha)$$
$${\bf \beta}^i = \mathrm{softmax} ({\bf W}_\beta(\tilde{\bf m}^i \circ {\bf h}^t_k) + b_\beta)$$
where ${\bf h}^t_k$ is the LSTM output for ${\bf u}^t_k$,  ${\bf W}_\alpha$ and ${\bf W}_\beta$ are weight matrices, $b_\alpha$ and $b_\beta$ are bias vectors, $\circ$ denotes element-wise vector multiplication, and $\mathrm{softmax}(x^i_j) = e^{x^i_j}\text{/}\sum_{k,l} e^{x^k_l}$.
Then we compute the weighted sum of memory entries:
$$\bar{\bf u} = \sum_{i,j} \alpha^i_j \tilde{\bf u}^i_j$$
$$\bar{\bf m} = \sum_i \beta^i \tilde{\bf m}^i$$

\subsection{Tag prediction} 
Finally we use $\bar{\bf u}$ and $\bar{\bf m}$ along with ${\bf h}^t_k$ to make the tag prediction:
$$\hat{\bf y}^t_{k,l} = \frac{\exp{{\bf z}^t_{k,l}}}{\sum_{l'} \exp{{\bf z}^t_{k,l'}}}$$
where ${\bf z}^t_k  =  {\bf W}_h {\bf h}_k^t +  {\bf W}_u \bar{\bf u} + {\bf W}_m \bar{\bf m}$ and ${\bf W}_*$ are all trainable weight matrices.

\section{Experiments}
\label{sec:experiments}

\begin{table}[t!]
\centering
\begin{tabular}{c||c|c|c}
\hline
Domain      & Train & Dev & Test \\ \hline
Orderfood    & 5,993 & 1,333 & 1,714 \\
Reservation & 5,130 & 1,215 & 1,339 \\
Taxi         & 6,120 & 1,216 & 1,414 \\
Events       & 4,347 & 822 & 850 \\
Movieticket & 3,781 & 712 & 738 \\ \hline
Average      & 5,074 & 1,059 & 1,211 \\ \hline
\end{tabular}
\caption{Data description}
\label{tab:data}
\end{table}

\begin{table*}[ht!]
\centering
\begin{tabular}{c||c|c|c|c|c|c}
\hline
Domain      & LSTM & +LS & +PreLS    & MNs & MNs+S & SSDMNs \\ \hline
Orderfood    & 87.38 & 93.70           & 94.10   & 92.07       & 94.36       & \textbf{96.45} \\
Reservation & 86.29 & 91.96          & 92.81   & 90.80      & 91.64       & \textbf{94.37} \\
Taxi        & 89.46 & 93.77          & 94.58  & 92.57       & 93.11       & \textbf{96.67} \\
Events      & 85.17 & 91.39          & 90.80   & 87.61       & 89.64       & \textbf{92.21} \\
Movieticket & 88.06 & 93.76          & 94.53   & 90.96      & 91.89       & \textbf{96.14} \\ \hline
Average     & 87.27 & 92.92          & 93.36  & 90.80       & 92.13       & \textbf{95.17} \\ \hline
\end{tabular}
\caption{Comparative performance in F1 score on five domains.}
\label{tab:nhot}
\end{table*}

\subsection{Models}
To evaluate the SSDMNs model, we conducted comparative experiments with the following systems:
\begin{itemize}
\setlength\itemsep{0pt}
\item \textbf{LSTM} uses the stateful LSTM only with no contextual information used.
\item \textbf{+LS} additionally uses the last system output.
\item \textbf{+PreLS} uses the pre-trained system tag projection matrix for encoding the last system output to show the efficacy of the CCA pre-training scheme.
\item \textbf{MNs} similarly to the Memory Networks model in Chen et al.~\cite{chen2016end}, the system does not use system's output information.
\item \textbf{MNs+S} MNs with system's output in surface form.
\end{itemize}

\subsection{Data}
The data is collected from five domains of Microsoft Cortana.
Table~\ref{tab:data} shows the number of utterances of the training, development and test datasets of each domain. All datasets were labeled by expert editors. 

\subsection{Training setting}
\label{subsec:setting}
In our experiments, all the models were implemented using Dynet \cite{neubig2017dynet} and were trained using Stochastic Gradient Descent (SGD) with Adam \cite{kingma2014adam}---an adaptive learning rate algorithm. We used the initial learning rate of $4 \times 10^{-4}$ and left all the other hyper parameters as suggested in \cite{kingma2014adam}. Each SGD update was computed with Intel MKL (Math Kernel Library) \footnote{https://software.intel.com/en-us/articles/intelr-mkl-and-c-template-libraries} without minibatching.
We used the dropout regularization~\cite{srivastava2014dropout} with the keep probability of 0.4. 
Both dimensions of the input and output of the stateful LSTM model were 100.
To initialize word embedding, we used the pre-trained GloVe model~\cite{pennington2014glove}. For system slot tag embedding, we used CCA to induce 100 dimensional vector representation from 20 personal assistant domains consisting of 3M utterances and 130 distinct tags. 


\subsection{Results}
\label{subsec:results}


To compute slot F1-score, we used the standard CoNLL evaluation script\footnote{http://www.cnts.ua.ac.be/conll2000/chunking/output.html}.
In Table~\ref{tab:nhot}, the stateful LSTM-only model (LSTM) achieved 87.27\% in averaged F1 score.
The improved performance of the +LS model, 92.92\%, confirms the same findings of previous studies on the effectiveness of previous turn information for contextual interpretation.
The further performance increase of the +PreLS model, achieving 93.36\%, demonstrates the possible gain with the proposed pre-training scheme.
Interestingly the MNs model couldn't show a better performance than the +LS model, even after incorporating system's output in surface form, justifying our novel contribution on the dual encoding of system output.
This claim is also supported by the best performance of the SSDMNs model reaching 95.17\%.

\section{Conclusion}
\label{sec:conclusion}
We presented a novel \emph{Speaker-Sensitive Dual Memory Networks} model for the multi-turn slot tagging task which extends the state-of-the-art Memory Networks-based approach with tailored memories to exploit the exact semantic knowledge on system output. We also described an efficient pre-training scheme to obtain better slot tag embeddings. The large performance improvement on real data demonstrates the efficacy of SSDMNs. Future work includes a detailed analysis on various reasoning behaviors of the model depending on different contexts. Also we plan to incorporate more elements of dialog state beyond system output. Another interesting direction is to perform multiple memory references to handle more complex discourse phenomena such as ellipsis, anaphora resolution.

\bibliography{acl2017}
\bibliographystyle{IEEEbib}

\end{document}